\documentclass[acmsmall,screen]{acmart}
\makeatletter

\newcommand{\Rmnum}[1]{\expandafter\@slowromancap\romannumeral #1@}
\makeatother
\newcommand{\tabincell}[2]{\begin{tabular}{@{}#1@{}}#2\end{tabular}} 
\usepackage{footnote}
\usepackage{multirow}
\usepackage{soul}
\usepackage{epstopdf}
\soulregister\cite7 
\soulregister\citep7 
\soulregister\citet7 
\soulregister\ref7 
\soulregister\pageref7 

\AtBeginDocument{%
  \providecommand\BibTeX{{%
    \normalfont B\kern-0.5em{\scshape i\kern-0.25em b}\kern-0.8em\TeX}}}

\setcopyright{acmcopyright}
\copyrightyear{2021}
\acmYear{2021}

\acmJournal{TOMM}

\begin{document}

\title{Decoupled Low-light Image Enhancement}

\thanks{This work was supported in part by the National Nature Science Foundation of China under Grant No. 62172137, 62072152, and 61725203, and in part by the Fundamental Research Funds for the Central Universities under Grant No. PA2020GDKC0023
and PA2019GDZC0095.}


\author{Shijie Hao}
\authornotemark[1]
\author{Xu Han}
\author{Yanrong Guo}
\author{Meng Wang}
\affiliation{%
  \institution{Key Laboratory
of Knowledge Engineering with Big Data (Hefei University of Technology), Ministry of Education and School of Computer Science and Information Engineering, Hefei University of Technology}
  \streetaddress{193 Tunxi Road}
  \city{Hefei}
  \country{China}}
\email{hfut.hsj@gmail.com}
\email{xuhan@mail.hfut.edu.cn}
\email{yrguo@hfut.edu.cn}
\email{eric.mengwang@gmail.com}

\renewcommand{\shortauthors}{S. Hao et al.}

\begin{abstract}
 The visual quality of photographs taken under imperfect lightness conditions can be degenerated by multiple factors, e.g., low lightness, imaging noise, color distortion and so on. Current low-light image enhancement models focus on the improvement of low lightness only, or simply deal with all the degeneration factors as a whole, therefore leading to a sub-optimal performance. In this paper, we propose to decouple the enhancement model into two sequential stages. The first stage focuses on improving the scene visibility based on a pixel-wise non-linear mapping. The second stage focuses on improving the appearance fidelity by suppressing the rest degeneration factors. The decoupled model facilitates the enhancement in two aspects. On the one hand, the whole low-light enhancement can be divided into two easier subtasks. The first one only aims to enhance the visibility. It also helps to bridge the large intensity gap between the low-light and normal-light images. In this way, the second subtask can be shaped as the local appearance adjustment. On the other hand, since the parameter matrix learned from the first stage is aware of the lightness distribution and the scene structure, it can be incorporated into the second stage as the complementary information. In the experiments, our model demonstrates the state-of-the-art performance in both qualitative and quantitative comparisons, compared with other low-light image enhancement models. In addition, the ablation studies also validate the effectiveness of our model in multiple aspects, such as model structure and loss function.  The trained model is available at~\url{https://github.com/hanxuhfut/Decoupled-Low-light-Image-Enhancement}.
\end{abstract}


\maketitle


\section{Introduction}
\label{section1}
The boom of smart phones and the Internet boosts the popularity of shooting and sharing photographs. However, the visual quality of photographs heavily suffers from the limited sensor size of phone cameras, when they are shot in darkness or imperfect lightness conditions. In addition, many low-quality low-light images have already been uploaded to the Internet, making themselves less attractive for collection or sharing. In this context, post-processing techniques for improving the visual quality of low-light images are highly desired.

The unsatisfactory visual quality of low-light images has two main aspects, that is, low scene visibility and poor appearance fidelity. The former one attributes to the low lightness during the imaging process, which lowers the pixel intensity and flattens the image contrast. As for the poor appearance fidelity, degeneration factors such as imaging noise, and color distortion result in unnatural visual effect of image appearance. Furthermore, the degeneration factors have negative impact on each other. For example, the low signal-noise-ratio (SNR) in the low-light regions makes it more difficult to remove noise. As for the color distortion, it is usually unnoticed in dark regions, as human vision is very limited in sensing color in darkness. Therefore, to obtain enhanced images with satisfying visibility and fidelity, the low-light enhancement model is expected to take all the degeneration factors into consideration.

However, many enhancement models overlook some degeneration factors in low-light images. For example, some enhancement models~\cite{reza2004realization,fu2016fusion,hao2019lightness} do not consider the noise issue, and inevitably produce amplified noise in the enhanced regions. The color distortion hidden in the darkness is also neglected in many low-light image enhancement models. For example, many Retinex-based models~\cite{cai2017joint,li2018structure} and fusion-based models~\cite{fu2016fusion,hao2019lightness} only enhance the intensity channel, or treat the three color channels independently. This strategy is incapable of correcting the color distortion, or even introduces extra color distortion into their enhanced results. There are also many novel methods trying to suppress the degeneration factors other than low lightness. For example, some models explicitly introduce the noise term into the Retinex-based image representation~\cite{li2018structure,TIP2020liu,hao2020low}. Despite of the clear physical meaning, these models have a common challenge of the ill-posed image decomposition, especially when additional constraints are introduced to suppress multiple degeneration factors. On the other hand, many enhancement models resort to end-to-end convolutional neural networks (CNNs), in which various loss terms are designed to comprehensively tackle the degeneration factors ~\cite{li2021lowlightSurvey}. These models treat the enhancement as the task of learning a pixel-wise mapping between image pairs. Since they often tackle all the degeneration factors as a whole, the mutual impacts between the degeneration factors are not fully explored, which may lead to a sub-optimal performance.

For the off-camera enhancement of low-light images, we can adopt the divide-and-conquer strategy to tackle the degeneration factors. We decouple the whole enhancing process into a two-stage enhancement, in which the first stage focuses on improving the scene visibility, while the second stage focuses on improving the appearance fidelity. The feasibility comes from the fact that the degeneration factors have different types of impacts on the image. As for the low lightness, the way it imposes on the scene is global and smooth. In the meanwhile, the lightness distribution of the imaging scene is jointly determined by the lightness strength and the scene structure. For example, even in a dim environment, the regions in the shade or not facing the light can be darker. So, the lightness distribution of low-light images has the regionally smooth property, which is in accordance with the Retinex theory ~\cite{li2014single,rother2011recovering}. As for the factors harming the appearance fidelity, the way they influence the appearance fidelity are much more local, especially for the imaging noise. This is fundamentally different with the characteristics of low-lightness. Based on the above observations, it is feasible to separate the suppression of low-lightness from the suppression of other degeneration factors by only introducing the local smoothness prior in the first stage.

\begin{figure*}[h]
\centering
\includegraphics[width=\linewidth]{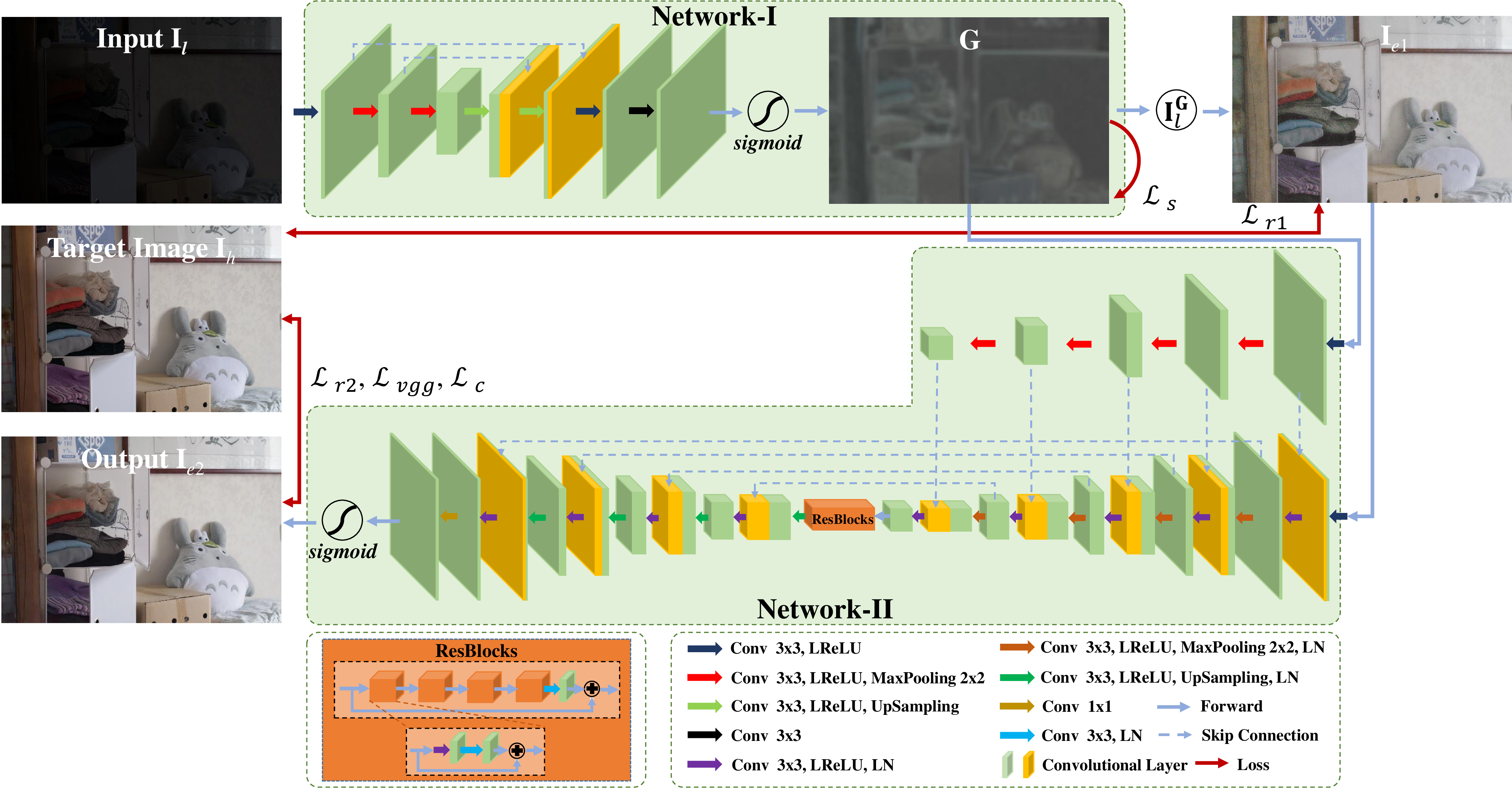}
\caption{The framework of the proposed model.}
\label{fig1}
\end{figure*}

The framework of our model is shown in Fig.~\ref{fig1}. The first stage of the model improves the visibility via a simple non-linear mapping, in which the pixel-wise parameters are learned by Network-I. At the second stage, Network-II is designed to learn the mapping function from the intermediately enhanced images produced at the first stage to the supervision images. The decoupling of the model facilitates the low-light enhancement in the following aspects. First, we can divide the comprehensive enhancement into two easier subtasks. Since the first subtask only aims to enhance the visibility, we can achieve this goal directly through the power function, instead of the ill-posed Retinex decomposition. As for the enhancement in the second stage, since the difference between the input and the supervision images is effectively shortened, the second subtask can be modeled as a process of learning to adjust image appearance under normal lightness. Second, the first-stage model provides some useful information that is helpful for the second subtask. Since we use the local smoothness prior in building Network-I, the learned parameter matrix is aware of both the scene lightness distribution and the scene structure. Therefore, in Network-II, we can introduce the matrix as the complementary information to guide the learning process. In the experiments, we evaluate our model on several public datasets, showing its effectiveness in various aspects and competitiveness against other state-of-the-art models.

The contributions of our research are summarized as follows. First, we propose a two-stage low-light image enhancement model, in which the whole enhancing process is decoupled into two sequential stages. Second, we build a network that learns a lightness-aware and structure-aware parameter matrix. Based on that, we can directly improve the visibility through the simple power function. This is fundamentally different from the traditional Retinex-based models and the deep-learning-based models following the Retinex framework. Third, under the guidance of the learned matrix, the network designed in the second stage learns to improve the visual quality by fully exploiting the fidelity between the pairwise training data.

The rest of the paper is organized as follows. Section~\ref{section2} briefly introduces the related works of low-light image enhancement. In Section~\ref{section3}, we describe the details of our model. Section~\ref{section4} reports the qualitative and quantitative experimental results and the analysis. Section~\ref{section5} finally concludes the paper.

\section{Related Work}
\label{section2}
In this section, we divide the low-light enhancement methods into the model-driven group and the data-driven group. In general, the first group aims to build models by carefully designing various enhancing rules or making full use of prior information. As for the second group, the data-driven models adopt neural networks to learn the mapping function with sufficient exemplar images. 

{\bf Model-driven Methods}. Histogram-based models~\cite{lee2013contrast,reza2004realization} expand the dynamic range of pixel intensities by reshaping the image histogram.The advantages of these methods lie in the simplicity and the fast speed. However, as they ignore the spatial information of an image, they are likely to produce undesirable visual effects, such as over- or under-enhancement.

 The Retinex-based models follow a common assumption according to the Retinex theory, i.e., an image can be represented by the product of an illumination layer describing the scene illumination distribution and a reflectance layer describing the object surface property. The enhancement can be firstly achieved by non-linearly mapping the illumination layer, and then multiplying it with the original reflectance layer. Therefore, these models boil down to an image decomposition problem in most cases. For example, Fu et al.~\cite{fu2016weighted} propose a weighted variational model to simultaneously decompose the illumination layer and the reflectance layer in the logarithmic domain. Cai et al.~\cite{cai2017joint} introduce prior information about illumination and texture into the objective function of the decomposition model, obtaining satisfying performances. In~\cite{li2018structure,ren2018joint,hao2020low,LR3M}, the imaging noise factor is considered in the model construction. ~\cite{hao2020low} tries to gradually suppress the noise during the decomposition process, while the others~\cite{li2018structure,ren2018joint} aim to explicitly extract the noise layer. In~\cite{LR3M}, Ren et al. propose a Low-Rank Regularized Retinex Model (LR3M) via imposing the low-rank prior on the objective function, aiming to suppress noise in the reflectance map. Different from the full Retinex models, simplified Retinex-based models~\cite{guo2016lime,zhang2018high,dong2011fast,sandoub2021low} estimate the illumination layer only, and treat the obtained reflection map as the final enhanced image. For example, Guo et al.~\cite{guo2016lime} roughly estimate the initial illumination map by extracting the maxRGB image from the color channels, and construct an edge-preserving filter to further refine the illumination map. As a post-processing step, BM3D is adopted to further denoise the obtained reflectance layer. Zhang et al.~\cite{zhang2018high} introduce prior information about color, structure and texture into the illumination map estimation, aiming at solving the issues of over-enhancement and artifacts caused by the simplified Retinex framework.

Fusion-based models also achieve good performance in the low-light image enhancement task. Ma et al.~\cite{ma2017robust} design a patch-based multi-exposure fusion model, which extracts and combines the best parts of the images taken under different exposures. However, the acquisition of multi-exposure images can sometimes be difficult, since a careful configuration of camera exposure parameters is needed. In addition, the pixel correspondence for the multiple-exposed images is non-negligible as the camera shaking or object motion is often encountered in the hand-held photographing. To solve this issue, with only one low-light image at hand, Fu et al.~\cite{fu2016fusion} generate several intermediately-enhanced images, and fuse them by using the Laplacian image pyramid. Similarly, based on the simplified Retinex model~\cite{guo2016lime}, Hao et al.~\cite{hao2019lightness} generate an intermediate enhanced image, and fuse it with the original image under the guidance of a structure- and illumination-aware weight map.

The common limitation of the model-driven methods mainly comes from the hand-crafted rules and various kinds of prior information, which tend to make the modeling and optimization process more and more complicated. Differently, our model adopts the data-driven paradigm that learns the function for mapping low-light images into normal-light images with higher visual quality. To some extent, the prior information about the degeneration factors is implicitly encoded into the model structure and the loss function in our model.

{\bf Data-driven Methods}. In recent years, deep neural networks have paved the way for the low-light image enhancement task~\cite{li2021lowlightSurvey,zhang2019kindling,lore2017llnet,wang2019underexposed,jiang2019enlightengan,guo2020zero,wei2018deep,yang2020fidelity,wang2018gladnet,li2021generative,li2021ambcr,tip2021}. According to the supervision level, the data-driven based enhancement models can be roughly divided into the supervised group, the semi-supervised group, and the unsupervised group. The supervised models rely on the pairwise low-light and normal-light images. Lore et al.~\cite{lore2017llnet} construct the seminal data-driven model LLNet for improving the image contrast via using a deep auto-encoder/decoder model. However, this model has the limitation of losing fine details. Thereafter, various CNN-based models have also been designed based on the Retinex decomposition framework. Wei et al.~\cite{wei2018deep} build two CNN-based networks for the Retinex decomposition and the following enhancement, respectively.  However, artifacts such as edge reversals can be introduced due to the inaccurate decomposed layers. Wang et al.~\cite{wang2019underexposed} bridge the gap between the under-exposure images and the skilled retouching images with an intermediate illumination map, thereby enhancing the network ability to adjust the photo illumination. 

As it is not easy to collect sufficient pairwise training data, many researchers resort to the unsupervised learning or the semi-supervised paradigm. As for the enhancement models based on unsupervised learning, Jiang et al.~\cite{jiang2019enlightengan} propose a GAN-based enhancement model by fully utilizing various global and local features extracted from unpaired low/normal-light images. Guo et al.~\cite{guo2020zero} propose the Zero-Reference Deep Curve Estimation (Zero-DCE) model, which formulates the enhancement as the image-specific curve estimation based on a deep network. By carefully constructing the loss function and the multi-exposure image datasets, this method performs well in enhancing the low lightness. However, these models are still limited in producing artifacts such as edge halo and color distortion. In~\cite{yang2020fidelity,tip2021}, Yang et al. propose a novel semi-supervised learning approach for low-light image enhancement. The model first uses paired low/normal-light images to restore signal through supervised learning, and then uses unpaired high-quality images to further enhance the perceptual quality of the image through unsupervised learning. These methods shed light on learning-based low-light image enhancement, as they relieve the burden of sufficient pairwise data. However, the issues such as stable training, color distortion, correlation of cross-domain information still remain open and challenging \cite{li2021lowlightSurvey}.

Our model belongs to the supervised group, but distinguishing itself in the decouple model structure that successive handles the degeneration factors. In addition, the model improves the visibility directly based on the power function, which is much easier than the Retinex decomposition. 

\section{Proposed Model}
\label{section3}
\subsection{Overview of the Proposed Model}
As shown in Fig.~\ref{fig1}, our model includes two stages. The first stage targets to solve the degeneration factor of the low-lightness in $\mathbf{I}_{l}$. Network-\Rmnum{1} is built to learn a parameter matrix $\mathbf{G}$, which is aware of both the illumination distribution and the scene structure. Based on the learned $\mathbf{G}$, the visibility of $\mathbf{I}_{l}$ can be adaptively adjusted through a pointwise non-linear mapping, producing the intermediate enhancement $\mathbf{I}_{e1}$. To some extent, $\mathbf{I}_{e1}$ can be seen as an image taken under normal lightness. In the second stage, Network-\Rmnum{2} is built to further enhance the visual quality of $\mathbf{I}_{e1}$ under the supervision of the appearance fidelity between its output $\mathbf{I}_{e2}$ and the target image $\mathbf{I}_{h}$. In this way, the degeneration factors in $\mathbf{I}_{e1}$, such as imaging noise and color distortion, can be effectively suppressed in this stage, and the lightness can be finely adjusted. The two stages are connected in a sequential way, that is, the output image $\mathbf{I}_{e1}$ of the first stage is taken as the input of Network-\Rmnum{2}. In addition, the learned matrix $\mathbf{G}$ from Network-\Rmnum{1} is also taken as the guidance in the encoder part in Network-\Rmnum{2} at multiple scales.

\subsection{Details of the Proposed Model}
\subsubsection{Network-\Rmnum{1} for Enhancing Visibility}
According to the Weber-Fechner Law, the human eye's perception of brightness is non-linear. Based on this, Retinex-based models improve the image lightness based on the following framework. First, the image $\mathbf{I}$ is decomposed into the illumination layer $\mathbf{I}_i$ and the reflectance layer $\mathbf{I}_r$, s.t. $\mathbf{I}=\mathbf{I}_i\odot\mathbf{I}_r$. Then, the non-linear function (or called Gamma correction) $\mathbf{I}_i^{g}$ is chosen to compensate for the defects of human vision in the darkness, where $g$ is empirically set as $1/2.2$. Finally, the enhanced image $\mathbf{I}_e$ is obtained via multiplying the two layers back, i.e., $\mathbf{I}_e=\mathbf{I}_i^{g}\odot\mathbf{I}_r$. In this way, the visibility is enhanced by the non-linear mapping of the decomposed layer $\mathbf{I}_i$. However, we note that the above process can be limited in two aspects. First, it is highly ill-posed to decompose $\mathbf{I}$ into $\mathbf{I}_i$ and $\mathbf{I}_r$. The quality of $\mathbf{I}_i$ and $\mathbf{I}_r$ is not guaranteed. Second, the parameter $g$ of the non-linear function is uniformly set as $1/2.2$ for all the pixels in $\mathbf{I}_i$, which is inadequate to handle the images with complex lightness conditions and contents.
\begin{figure*}[htbp]
\centering
\includegraphics[width=\linewidth]{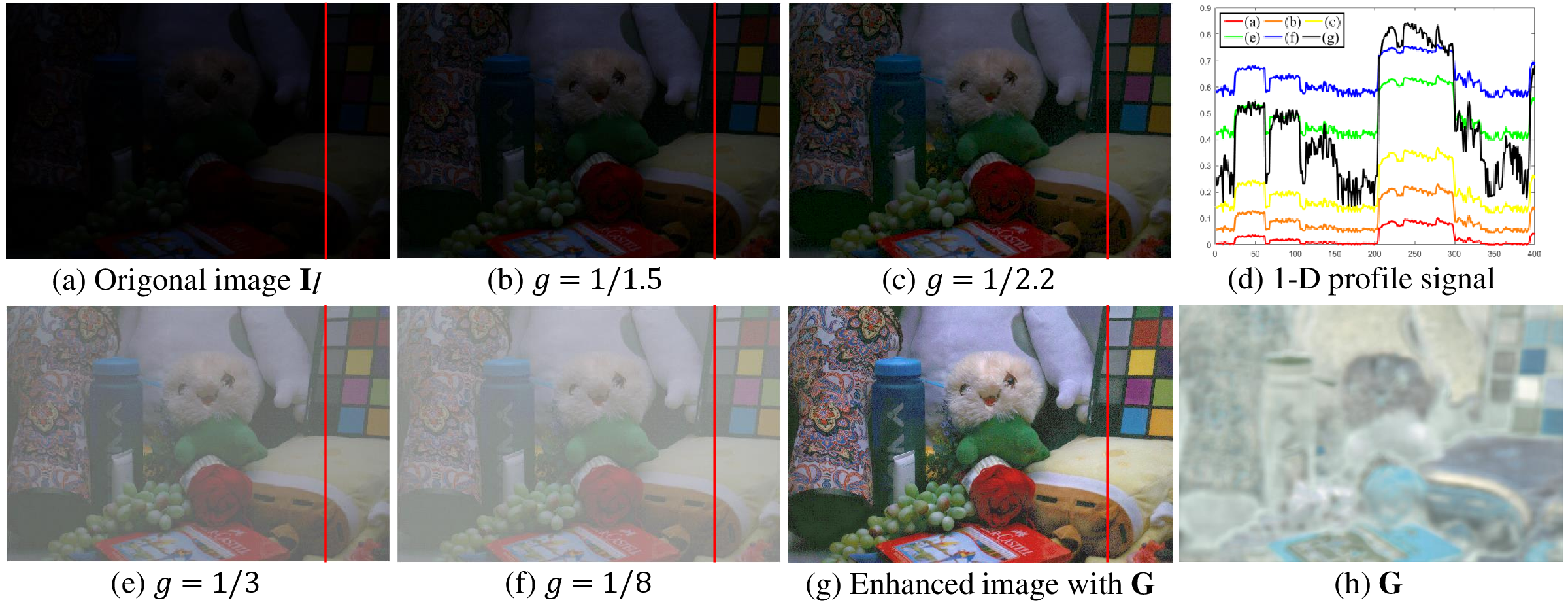}
\caption{The results of illumination improvement using the uniform or pixel-wise non-linear mapping.}
\label{fig2}
\end{figure*}

To solve these problems, we simplify the above process by using the power function $y=x^{g}$ only. Instead of a pre-defined global parameter $g$, we try to learn different $g$ values for the pixels in $\mathbf{I}_{l}$. The feasibility of this simplification comes from the fact that the non-linear power function $y=x^{g}$ ($x\in(0,1]$) is able to map pixel values to different levels. In Fig.~\ref{fig2}, we show the 1D intensity profiles along the red lines of (a-c, e-f) in (d) as an example. We can see that the pixel values of (a) can be lifted with different $g$ values. On the other hand, however, it is insufficient to only improve the overall lightness with a single empirically-set $g$ value. As shown in Fig.~\ref{fig2} (b, c, e, f), we can see that a large $g$ (e.g. $1/1.5$) leads to under-enhancement, while a small $g$ (e.g. $1/8$) leads to over-enhancement. The contrast in (b, c, e, f) is still very poor. In this context, a parameter matrix $\mathbf{G}$ is learned to adaptively improve the visibility via the point-wise power function $\mathbf{I}_{e1}=\mathbf{I}_{l}(m,n)^{\mathbf{G}(m,n)}$. As we can see in Fig.~\ref{fig2}, the enhanced image (g) based on the learned $\mathbf{G}$ is far better than (b, c, e, f) in terms of the image contrast. This can be also validated by taking a closer look at the 1D profiles shown in Fig.~\ref{fig2} (d). Obviously, the 1D profile based on our model (in black color) performs best in terms of extending the contrast between the adjacent content regions. As for the reason, since the learned $\mathbf{G}$ is consistent with the content structure of $\mathbf{I}_{l}$, the different regions in $\mathbf{I}_{l}$ receive their suitable enhancing strengths. In the following, we introduce the details of Network-\Rmnum{1}.  

We choose a five-layer lightweight Unet structure~\cite{ronneberger2015u} as the backbone of Network-\Rmnum{1}. As shown in Fig.~\ref{fig1}, it is mainly composed of $3\times 3$ convolutional layers, up/down sampling layers, and LReLU~\cite{he2015delving} activation functions. At the end of the network, a sigmoid function is imposed on $\mathbf{G}$ to map all its values into $\left[ 0,1 \right]$. 

Denote $\mathbf{I}_l$ and $\mathbf{I}_h$ as a pair of low-light image and normal-light image with pixel-wise correspondence, we first design the loss term $\mathcal{L}_{r1}$ to describe the lightness difference between the enhanced image and the ground truth image:
\begin{equation}
\mathcal{L}_{r1}=\lVert \mathbf{I}_{l} ^{\mathbf{G}}-\mathbf{I}_{h} \rVert ^{2}_{2}.
\label{eq1}
\end{equation}
Then, we impose a regularization term of $\mathbf{G}$ to constrain its local smoothness, as shown in Eq.~\ref{eq2}:
\begin{equation}
\mathcal{L}_{s}=\frac{1}{3}\sum_{c=\left\{ R,G,B \right\}}{\lVert \left| \nabla _x\mathbf{G}_c \right|+\left| \nabla _y\mathbf{G}_c \right| \rVert}^{2}_{2},
\label{eq2}
\end{equation}
where $\mathbf{G}_c$ represents one of the three channels of $\mathbf{G}$, and $\nabla _{x}$ or $\nabla _{y}$ indicates the differential of $\mathbf{G}_c$ in the $x$ or $y$ direction. As stated in the first section, the consideration of introducing this loss term comes from the local smoothness prior that the lightness distribution is always locally smooth. Therefore, the local smoothness in $\mathbf{G}$ helps to preserve the monotonicity relations between neighboring pixels. In addition, it helps to avoid the over-fitting during training Network-\Rmnum{1}.

Based on Eq.~\ref{eq1} and Eq.~\ref{eq2}, the total loss function of Network-\Rmnum{1} can be expressed as 
\begin{equation}
\mathcal{L}_{total1}=\mathcal{L}_{r1}+w_s \mathcal{L}_{s},
\label{eq3}
\end{equation}
where $w_s$ is the balancing weight. 

With the learned $\mathbf{G}$, the output image of the first stage $\mathbf{I}_{e1}$ is obtained via the pixel-wise mapping $\mathbf{I}(m,n)^{\mathbf{G}(m,n)}$. From the above formulation, we can see that the enhancement at this stage focuses on dealing with the lightness.

\begin{figure*}[htbp]
\centering
\includegraphics[width=\linewidth]{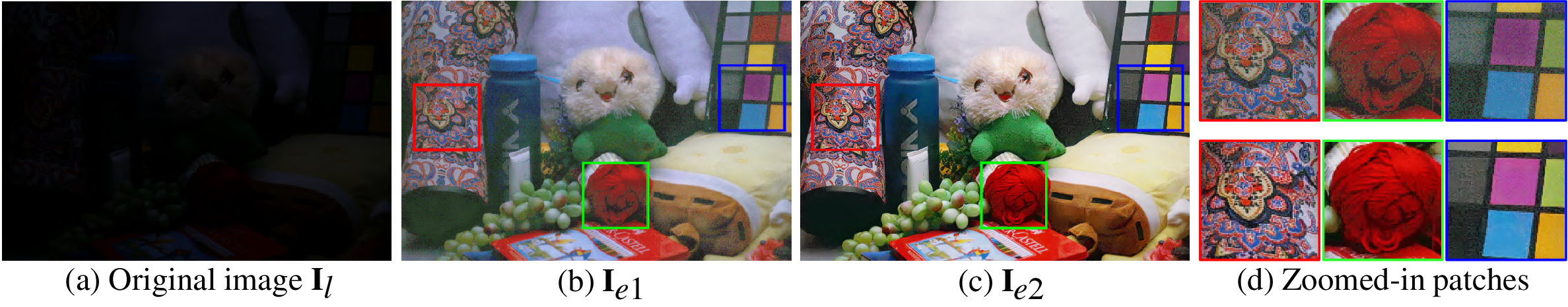}
\caption{Comparison of results from the first stage and the whole model.}
\label{fig3}
\end{figure*}

\subsubsection{Network-\Rmnum{2} for Preserving Appearance Fidelity}
Although the first stage of our model enhances the poor lightness in $\mathbf{I}_{l}$, other degeneration factors introduced in the image acquisition still exist in $\mathbf{I}_{e1}$, such as the remaining imaging noise and color distortion. In this context, the second stage of our model is designed to further improve the visual quality of $\mathbf{I}_{e1}$. From the example shown in Fig.~\ref{fig3}, we compare the quality of $\mathbf{I}_{e1}$ and the final result $\mathbf{I}_{e2}$. From the zoomed-in patches in (d), it is obvious that the imaging noise has been suppressed, and the color distortion has also been corrected in $\mathbf{I}_{e2}$. In this way, the appearance fidelity is better preserved by Network-\Rmnum{2}.

As for the Network-\Rmnum{2}'s structure, we adopt the ResUnet that incorporates four residual blocks between the encoder and the decoder. The skip connections and the residual blocks facilitate the utilization of the low-level details during the image reconstruction. Of note, we expand the network structure by making full use of $\mathbf{G}$ learned from the first stage. We also build an encoder structure for $\mathbf{G}$ that has the same dimension with the ResUnet's encoder part, aiming at concatenating $\mathbf{G}$ and $\mathbf{I}_{e1}$ at different scales. In this way, the hint of the structure-aware lightness distribution in $\mathbf{G}$ can be used as the guidance at the second stage. The details of Network-\Rmnum{2}'s structure can be seen in Fig.~\ref{fig1}, which is mainly composed of $3\times 3$ convolutional layers, up/down sampling layers, LReLU activation functions, and Layer Normalizations~\cite{ba2016layer}. Finally, a sigmoid function is added to normalize the values in $\mathbf{I}_{e2}$ into $[0,1]$. It is noted that we attempt to build Network-\Rmnum{2} with a much larger model size than Network-\Rmnum{1}, as it aims at adjusting the local image appearance with subtle changes.  

The loss function of Network-\Rmnum{2} is designed to seek the fidelity between the Network-\Rmnum{2}'s output $\mathbf{I}_{e2}$ and the ground truth normal-light image $\mathbf{I}_h$. The first term $\mathcal{L}_{r2}$ keeps the fidelity at the pixel level:
\begin{equation}
\mathcal{L}_{r2}=\lVert \mathbf{I}_{e2} - \mathbf{I}_{h}  \rVert ^{2}_{2}.
\label{eq4}
\end{equation}
In addition to the first stage, $\mathcal{L}_{r2}$ further provides the chance to finely adjust the illumination of $\mathbf{I}_{e1}$.

To better suppress the imaging noise, we also introduce the VGG-based loss term based on the pre-trained VGG-19 network~\cite{simonyan2014very}. The VGG loss~\cite{ledig2017photo} can be used to measure the feature similarity between $\mathbf{I}_{e2}$ and $\mathbf{I}_h$ at a higher level:
\begin{equation}
\mathcal{L}_{vgg}= \frac{1}{C_iH_iW_i} \lVert \psi_i(\mathbf{I}_{e2}) - \psi_i(\mathbf{I}_h) \rVert,
\label{eq5}
\end{equation}
where $\psi_i(*)$ is the feature map obtained by the $i$-th convolutional layer of the VGG-19 network, and $W_i$, $H_i$ and $C_i$ denote the feature map dimension. This loss is based on the feature produced by the $relu\_5\_4$ layer of VGG-19, which is aware of scene semantics. Different from $\mathcal{L}_{r2}$ focusing on the difference between two image pixels, $\mathcal{L}_{vgg}$ drives the network to pay more attentions to the content consistency. 

As discussed in Section~\ref{section1}, the human vision system becomes more sensitive to colors under a lightened environment. Therefore, the appearing color distortion in $\mathbf{I}_{e1}$ should be corrected at this stage. To this end, we introduce a loss term that targets at correcting the color distortion in $\mathbf{I}_{e1}$. Inspired by~\cite{hu2017fc}, we aim to minimize the angle of the color vectors between $\mathbf{I}_{e2}$ and $\mathbf{I}_h$: 
\begin{equation}
\mathcal{L}_{c}=\frac{180}{\pi}\mathrm{arccos}(\frac{\mathbf{I}_{e2}}{\lVert \mathbf{I}_{e2} \rVert _{2}}\cdot\frac{\mathbf{I}_h}{\lVert \mathbf{I}_h \rVert _{2}}).
\label{eq6}
\end{equation}

In Eq.~\ref{eq6}, each image pixel is represented as a normalized RGB vector, and the color consistency is measured by the angle between the two vectors. 

Therefore, the total loss function of Network-\Rmnum{2} can be expressed as
\begin{equation}
\mathcal{L}_{total2}=\mathcal{L}_{r2}+w_{vgg} \mathcal{L}_{vgg}+w_{c} \mathcal{L}_{c},
\label{eq7}
\end{equation}
where $w_{vgg}$ and $w_{c}$ are the weights for the $\mathcal{L}_{vgg}$ and $\mathcal{L}_{c}$, respectively. From the above formulation, we can see that Network-\Rmnum{2} plays the role of fine tuning the image appearance of $\mathbf{I}_{e1}$, therefore suppressing the degeneration factors that harm the appearance fidelity.

\section{Experiments}
\label{section4}
In this section, we first describe the implementation details. Then, we present and analyze the experimental results, including the qualitative and quantitative comparison with other low-light enhancement models, and the ablation studies of our model.

\subsection{Implementation Details}
We use the LOL dataset and parts of the synthetic low/normal-light images from~\cite{wei2018deep} to train our model. The LOL dataset includes 500 low/normal-light natural image pairs. We divide them into three parts, that is, 477 pairs as the training set, 8 pairs as the validation set, and 15 pairs as the testing set. As for the synthetic data, we choose the first 100 pairs from the whole set as a part of training data. In addition, we perform data augmentation by rotating and flipping the images. Since the proposed model is composed of two stages, the training process is divided into two phases as well. In the first phase, we use the LOL dataset set as the training dataset for Network-\Rmnum{1}. We set the batch size as 10, the patch size as $48\times48$, and the epoch as 2000. In the second phase, we use the LOL dataset and the 100 pairs of synthetic images as the training dataset for Network-\Rmnum{2}. We set the batch size as 8, the patch size as $256\times256$, and the epoch as 1000. In both phases, the networks are optimized by the ADAM optimizer, and the initial learning rate is set as 0.0001. As for the weights in Eq.~\ref{eq3} and Eq.~\ref{eq7}, $w_{s}$, $w_{vgg}$ and $w_{c}$ are set as 20, 1, 0.2, respectively. The entire model is trained on an NVIDIA GTX 2080Ti GPU and an Intel(R) Xeon(R) Silver 2.10GHz CPU using the Tensorflow framework.

\subsection{Performance Evaluation}
\subsubsection{Qualitative and Quantitative Comparison with Other Models}
We choose eleven models for comparison, including HQEC~\cite{zhang2018high}, LIME~\cite{guo2016lime} (with the steps of post-denoising and fusion), MF~\cite{fu2016fusion}, RRM~\cite{li2018structure}, Retinex-Net~\cite{wei2018deep}, KinD~\cite{zhang2019kindling}, DeepUPE~\cite{wang2019underexposed}, SICE~\cite{cai2018learning}, Zero-DCE~\cite{guo2020zero}, DRBN~\cite{yang2020fidelity}, EnlightenGAN~\cite{jiang2019enlightengan}. Among them, the first four models are model-driven ones, and the rest ones belong to the data-driven group. Of note, the training dataset of the DRBN~\cite{yang2020fidelity} model contains data from the LOL testing dataset~\cite{wei2018deep}. For fairness, we retrain this model according to our dataset configuration, while keeping all the other implementation details in ~\cite{yang2020fidelity} unchanged. We evaluate all the models on the LOL testing dataset~\cite{wei2018deep}, as well as four public datasets without ground truth images, i.e. DICM~\cite{lee2012contrast}, LIME~\cite{guo2016lime}, Fusion~\cite{wang2016fusion} and VV~\cite{VV}. The total number of images for testing is 15+44+10+18+24=111. In particular, to facilitate the implementation of all the methods for comparison, the images in the VV dataset are resized. 
\begin{figure*}[htbp]
\centering
\includegraphics[width=\linewidth]{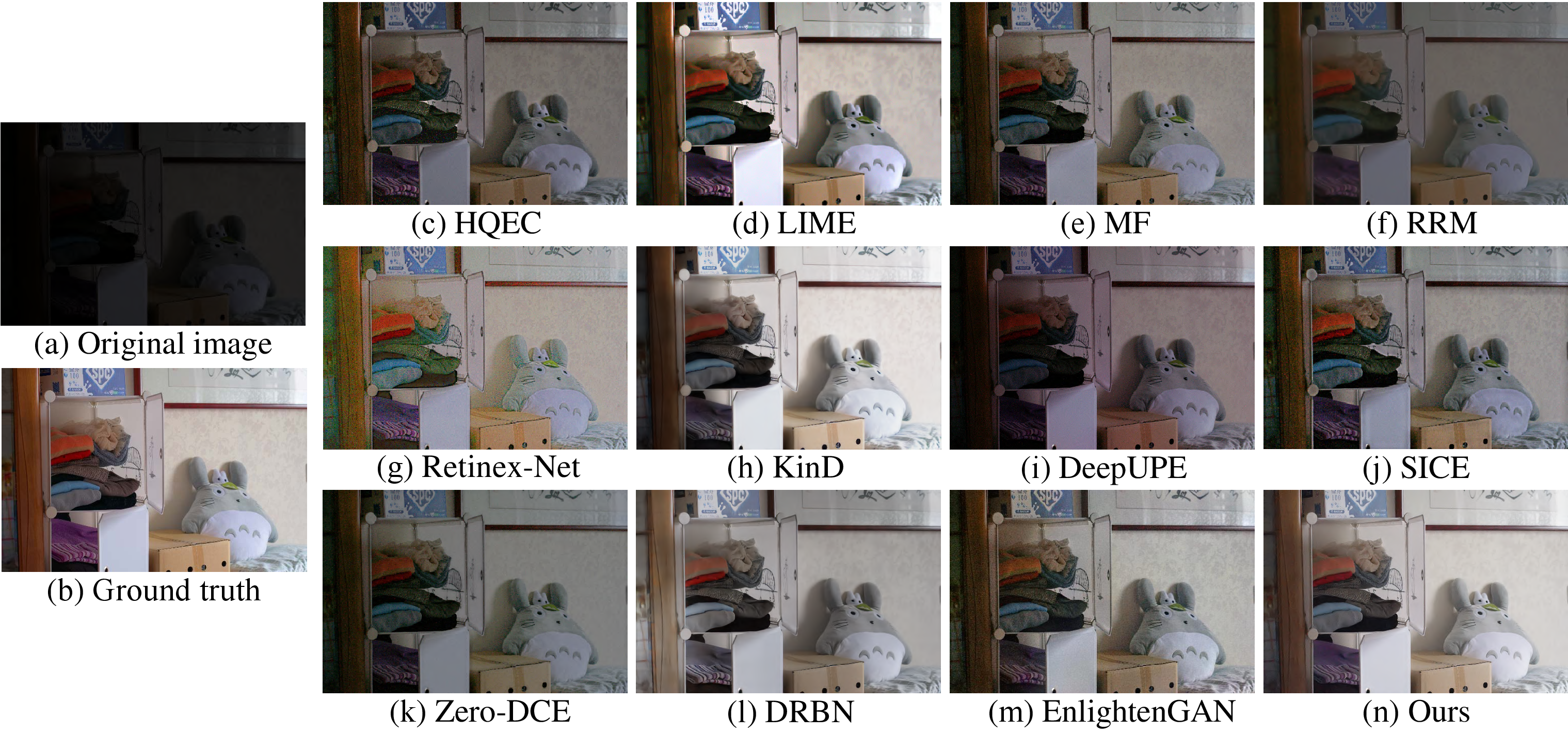}
\caption{Comparison with eleven low-light image enhancement methods. (a) Original image. (b) Ground truth. (c) HQEC~\cite{zhang2018high}. (d) LIME~\cite{guo2016lime}. (e) MF~\cite{fu2016fusion}. (f) RRM~\cite{li2018structure}. (g) Retinex-Net~\cite{wei2018deep}. (h) KinD~\cite{zhang2019kindling}. (i) DeepUPE~\cite{wang2019underexposed}. (j) SICE~\cite{cai2018learning}. (k) Zero-DCE~\cite{guo2020zero}. (l) DRBN~\cite{yang2020fidelity}. (m) EnlightenGAN~\cite{jiang2019enlightengan}. (n) Ours. It is better to check these results in a zoomed-in view (same for the following figures).}
\label{fig4}
\end{figure*}
\begin{figure*}[htbp]
\centering
\includegraphics[width=\linewidth]{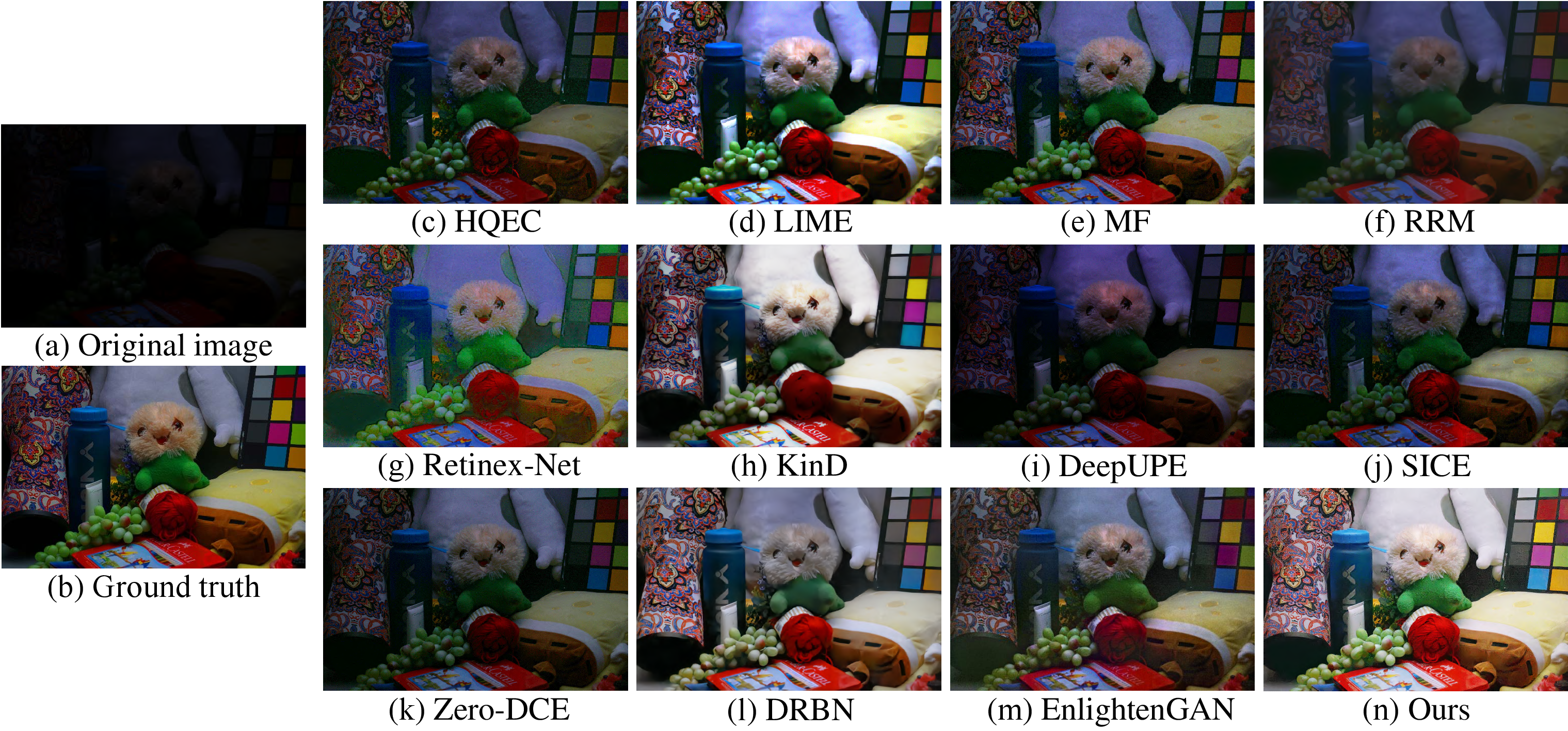}
\caption{Comparison with eleven low-light image enhancement methods. (a) Original image. (b) Ground truth. (c) HQEC~\cite{zhang2018high}. (d) LIME~\cite{guo2016lime}. (e) MF~\cite{fu2016fusion}. (f) RRM~\cite{li2018structure}. (g) Retinex-Net~\cite{wei2018deep}. (h) KinD~\cite{zhang2019kindling}. (i) DeepUPE~\cite{wang2019underexposed}. (j) SICE~\cite{cai2018learning}. (k) Zero-DCE~\cite{guo2020zero}. (l) DRBN~\cite{yang2020fidelity}. (m) EnlightenGAN~\cite{jiang2019enlightengan}. (n) Ours.}
\label{fig5}
\end{figure*}
\begin{figure*}[htbp]
\centering
\includegraphics[width=\linewidth]{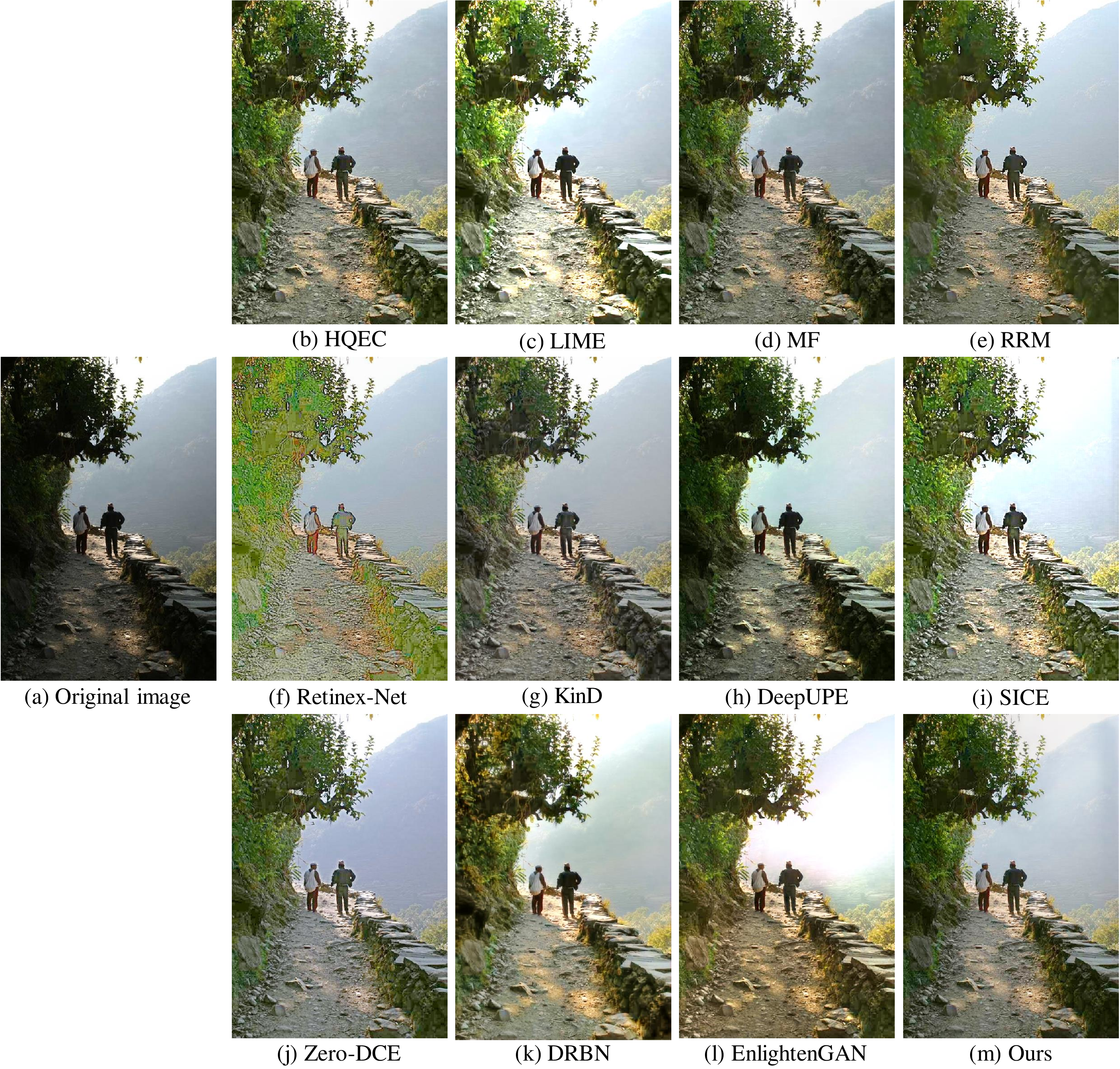}
\caption{Comparison with eleven low-light image enhancement methods. (a) Original image. (b) HQEC~\cite{zhang2018high}. (c) LIME~\cite{guo2016lime}. (d) MF~\cite{fu2016fusion}. (e) RRM~\cite{li2018structure}. (f) Retinex-Net~\cite{wei2018deep}. (g) KinD~\cite{zhang2019kindling}. (h) DeepUPE~\cite{wang2019underexposed}. (i) SICE~\cite{cai2018learning}. (j) Zero-DCE~\cite{guo2020zero}. (k) DRBN~\cite{yang2020fidelity}. (l) EnlightenGAN~\cite{jiang2019enlightengan}. (m) Ours.}
\label{fig6}
\end{figure*}
\begin{figure*}[htbp]
\centering
\includegraphics[width=\linewidth]{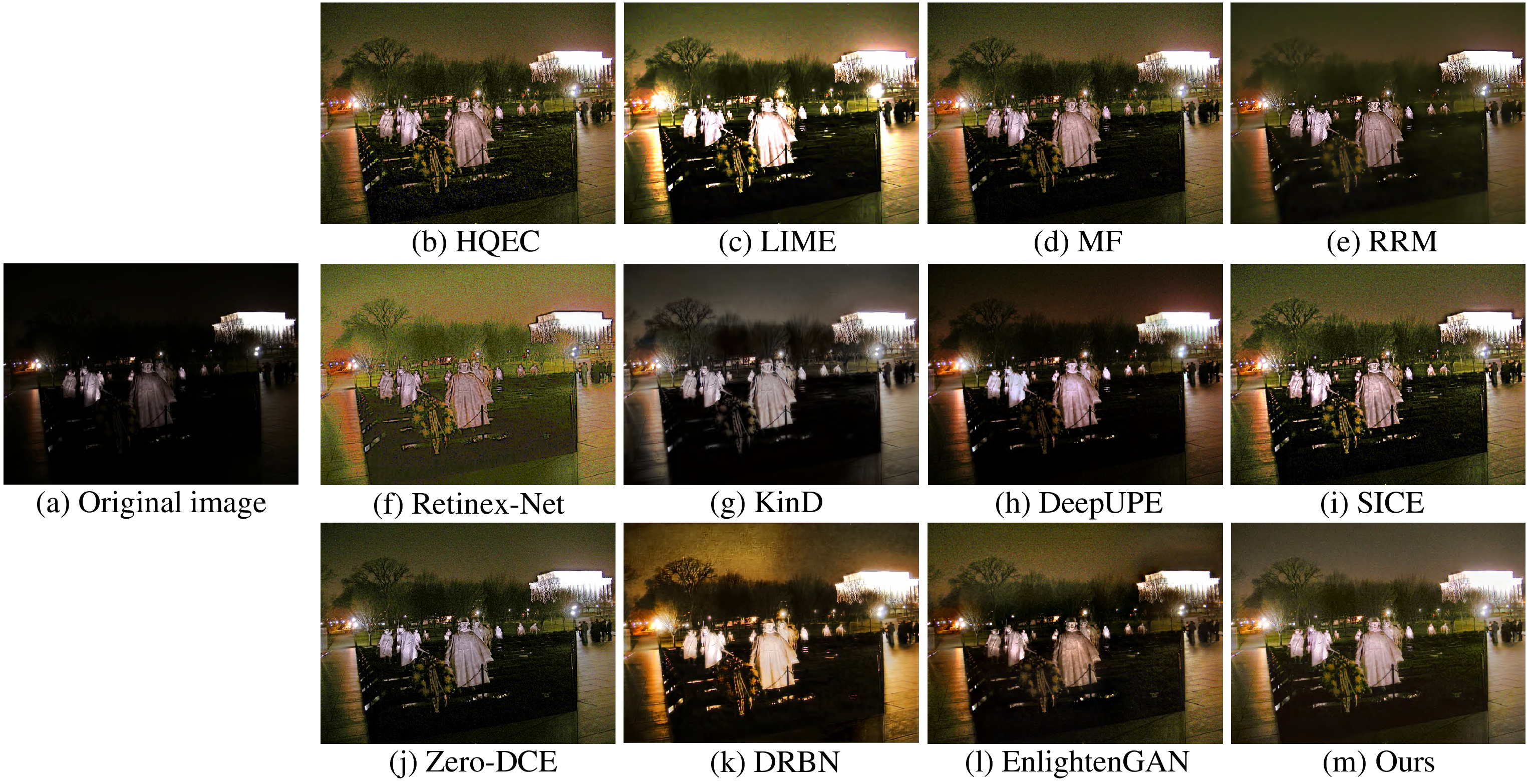}
\caption{Comparison with eleven  low-light image enhancement methods. (a) Original image. (b) HQEC~\cite{zhang2018high}. (c) LIME~\cite{guo2016lime}. (d) MF~\cite{fu2016fusion}. (e) RRM~\cite{li2018structure}. (f) Retinex-Net~\cite{wei2018deep}. (g) KinD~\cite{zhang2019kindling}. (h) DeepUPE~\cite{wang2019underexposed}. (i) SICE~\cite{cai2018learning}. (j) Zero-DCE~\cite{guo2020zero}. (k) DRBN~\cite{yang2020fidelity}. (l) EnlightenGAN~\cite{jiang2019enlightengan}. (m) Ours.}
\label{fig7}
\end{figure*}

We conduct visual comparisons between the results of our model and the others in Figs.~\ref{fig4}-\ref{fig7}. Among them, Figs.~\ref{fig4} and~\ref{fig5} are examples from the LOL testing dataset~\cite{wei2018deep}, while Figs.~\ref{fig6} and~\ref{fig7} are from the DICM dataset~\cite{lee2012contrast}. We have some observations from Figs.~\ref{fig4} and~\ref{fig5}. As for the model-driven ones, HQEC~\cite{zhang2018high}, MF~\cite{fu2016fusion}, and RRM~\cite{li2018structure} fail to improve the lightness into a satisfying level. LIME~\cite{guo2016lime} improves the overall scene lightness, but still leaves locally unclear regions. In addition, despite the usage of some post-processing steps for denoising and preserving details, the color distortion still exists in its results, showing the necessity of the comprehensive enhancement for low-light images. As for the data-driven models, the results of Retinex-Net~\cite{wei2018deep} and SICE~\cite{cai2018learning} have strong noise and artifacts that degrade their visual quality. The similar trouble occurs in the result of EnlightenGAN~\cite{jiang2019enlightengan} to a certain extent. The enhancement results of 
DRBN~\cite{yang2020fidelity} show uneven enhancement, such as bright spots or shadows (The similar problem occurs in Figs.~\ref{fig6} and~\ref{fig7}). The results of DeepUPE~\cite{wang2019underexposed} and Zero-DCE~\cite{guo2020zero} demonstrate insufficiently enhanced illumination, making many details still unclear. The results of KinD~\cite{zhang2019kindling} have some locally under-enhanced regions, as well as distorted colors. Our results obtain better visual quality than others in terms of global visibility and local appearance. As for the images from datasets other than LOL, satisfying performance can still be observed in Figs.~\ref{fig6} and~\ref{fig7}. In Figs.~\ref{fig6}, compared with other models, our model effectively lightens the dark regions to a moderate level, while 
preserving the visual naturalness of the result. In Fig.~\ref{fig7}, apart from improving the overall lightness, our model performs best in suppressing the imaging noise hidden in the originally dark regions (e.g. the sky region) while preserving the small-scale details (e.g. the trees). 
\begin{table*}[htbp]
\caption{Comparison of PSNR, SSIM~\cite{wang2004image} and MS-SSIM scores of different methods}
\label{tab1}
\small
\tabcolsep 8pt 
\begin{tabular}{c|c}
\toprule
   Methods & LOL~\cite{wei2018deep}\\\hline
   MF~\cite{fu2016fusion} & 16.9962/0.6049/0.9058  \\\hline
   
    HQEC~\cite{zhang2018high}  &  14.5775/0.5903/0.8906 \\\hline
   
   LIME~\cite{guo2016lime} &  17.0172/0.6062/0.8731    \\\hline
   
   RRM~\cite{li2018structure} &  13.8765/0.6577/0.8320  \\\hline
   
    Retinex-Net~\cite{wei2018deep}&16.7740/0.5594/0.8176\\\hline
   
   KinD~\cite{zhang2019kindling}& \underline{20.3792}/\underline{0.8045}/0.9365 \\\hline
   
   DeepUPE~\cite{wang2019underexposed}&  12.2068/0.4783/0.8251  \\\hline
   
   
   SICE~\cite{cai2018learning}& 16.0889/0.5801/0.8731 \\\hline
   
   Zero-DCE~\cite{guo2020zero}&  14.8607/0.5849/0.8955  \\\hline
   
   DRBN~\cite{yang2020fidelity}&  19.7475/0.7961/\underline{0.9543}\\\hline
   
   EnlightenGAN~\cite{jiang2019enlightengan}&  17.4866/0.6661/0.9072\\\hline
   
   Ours&  {\bf 21.8382 }/{\bf 0.8216}/{\bf 0.9606}\\
\bottomrule
\end{tabular}
\\\footnotesize{The best and second-best results are highlighted with {\bf bold} \\and \underline{underline}, respectively (same for the following tables).}\\
\end{table*}
\begin{table*}[htbp]
\caption{Comparison of  NIQE~\cite{mittal2012making}, PIQE~\cite{venkatanath2015blind}, BRISQUE~\cite{mittal2011blind} and NFERM~\cite{gu2014using} scores of different methods}
\label{tab2}
\small
\tabcolsep 5pt 
\begin{tabular*}{\textwidth}{c|c|c|c|c|c|c|c}
\toprule
   Methods & Metrics & LOL~\cite{wei2018deep} & DICM~\cite{lee2012contrast} & LIME~\cite{guo2016lime} & VV~\cite{VV} &  Fusion~\cite{wang2016fusion} & Average\\\hline
   MF~\cite{fu2016fusion}  & \tabincell{c}{ NIQE\\ PIQE\\ BRISQUE \\ NFERM} & \tabincell{c}{9.7125\\47.1974\\34.7835\\35.0411}&\tabincell{c}{3.4533\\33.9031\\22.9418\\ 15.5248}&\tabincell{c}{4.1025\\36.3805\\22.3231\\13.3566 }&\tabincell{c}{2.4371\\27.3274\\18.9399\\13.7342}&\tabincell{c}{\underline{2.8842}\\35.7042\\\underline{22.6970}\\18.1297}&\tabincell{c}{4.0456\\34.7931\\23.5813\\18.0021}\\\hline
   
    HQEC~\cite{zhang2018high}&\tabincell{c}{ NIQE\\ PIQE\\ BRISQUE \\ NFERM} &\tabincell{c}{8.7359\\41.4826\\32.6709\\30.2939 }&\tabincell{c}{3.6758\\36.9067\\23.7496\\18.4524}  &\tabincell{c}{4.2784\\36.6805\\22.0326\\13.7183}&\tabincell{c}{2.5375\\27.2834\\19.0154\\13.4168}&	\tabincell{c}{2.9474\\35.3115\\23.3774\\\underline{4.7154}} &\tabincell{c}{4.0496\\35.1653\\23.7165\\16.3097}   \\\hline
   
   LIME~\cite{guo2016lime}&\tabincell{c}{ NIQE\\ PIQE\\ BRISQUE \\ NFERM}&\tabincell{c}{4.1844\\43.9761\\26.7223\\ \underline{ 9.1808}}&\tabincell{c}{2.8592\\44.3125\\24.0232\\14.1868}&\tabincell{c}{4.1842\\51.3510\\30.0871\\31.2363} &\tabincell{c}{2.4030\\46.1212\\24.7279\\17.8949} &	\tabincell{c}{3.3068\\50.6316\\29.2682\\27.8328}  &\tabincell{c}{3.1212\\46.3169\\25.9371\\18.0609}    \\\hline
   
   RRM~\cite{li2018structure}&\tabincell{c}{ NIQE\\ PIQE\\ BRISQUE \\ NFERM}&\tabincell{c}{3.9517\\47.3000\\34.9902\\19.7309} &	\tabincell{c}{3.3186\\42.5862\\28.6429\\11.5682}& \tabincell{c}{4.1080\\42.6625\\30.2665\\22.5246}&\tabincell{c}{2.7928\\39.0783\\26.1066\\12.9621} &\tabincell{c}{3.2426\\46.3104\\29.5513\\19.9871} &\tabincell{c}{3.3953\\43.0195\\29.2307\\15.1719}  \\\hline
   
    Retinex-Net~\cite{wei2018deep}&\tabincell{c}{ NIQE\\ PIQE\\ BRISQUE \\ NFERM}& \tabincell{c}{9.7297\\57.6731\\39.5860\\40.2884}& \tabincell{c}{4.7121\\40.9089\\26.6334\\19.3010}& \tabincell{c}{4.9079\\42.7741\\26.1007\\22.0788}& \tabincell{c}{3.2440\\30.3951\\21.2306\\23.2519}&\tabincell{c}{3.6738\\39.5889\\24.8595\\7.5005}& \tabincell{c}{4.9220\\40.8551\\26.8799\\21.3280}\\\hline
   
   KinD~\cite{zhang2019kindling}&\tabincell{c}{ NIQE\\ PIQE\\ BRISQUE \\ NFERM}&\tabincell{c}{3.9849\\65.6035\\32.4365\\24.5540}&\tabincell{c}{2.9915\\45.6465\\28.3032\\17.0659 }&\tabincell{c}{4.3609\\45.3092\\26.7728\\34.8583}&\tabincell{c}{\underline{2.2350}\\37.7389\\23.5457\\14.1242}&\tabincell{c}{2.9653\\43.3237\\25.2592\\19.4639}&\tabincell{c}{3.0813\\46.2266\\27.2016\\19.4335}\\\hline
   
   DeepUPE~\cite{wang2019underexposed}&\tabincell{c}{ NIQE\\ PIQE\\ BRISQUE \\ NFERM}& \tabincell{c}{7.9474\\24.5520\\29.3246\\23.7664}&\tabincell{c}{3.2082\\29.9233\\\underline{21.8756}\\7.9156}& \tabincell{c}{\underline{3.5689}\\34.7332\\24.3930\\17.7925}& \tabincell{c}{2.2652\\27.5227\\18.5365\\\underline{11.9283}}& \tabincell{c}{2.8900\\37.0041\\{\bf 21.3188}\\18.5893}& \tabincell{c}{3.6256\\30.2600\\22.2968\\13.5459}\\\hline
   
   
   SICE~\cite{cai2018learning}&\tabincell{c}{ NIQE\\ PIQE\\ BRISQUE \\ NFERM}&\tabincell{c}{9.0110\\46.0741\\34.0779\\33.8036}&\tabincell{c}{4.0834\\36.3870\\28.1778\\17.3339}& \tabincell{c}{4.2229\\34.5711\\20.8475\\	{\bf 11.0754}}& \tabincell{c}{3.0799\\29.5486\\26.2183\\14.5109}& \tabincell{c}{3.8690\\{\bf28.6528}\\25.2124\\{\bf 2.5557}}& \tabincell{c}{4.5101\\34.7997\\27.4102\\15.9889}\\\hline
   
   Zero-DCE~\cite{guo2020zero}&\tabincell{c}{ NIQE\\ PIQE\\ BRISQUE \\ NFERM}&\tabincell{c}{8.2233\\35.5059\\30.3051\\26.6630} &\tabincell{c}{{\bf2.6963}\\\underline{25.9288}\\25.1439\\13.9139}&\tabincell{c}{3.7891\\35.8671\\23.3341\\\underline{11.5189}}&\tabincell{c}{2.5787\\29.6227\\24.6320\\15.3613}&\tabincell{c}{3.2615\\36.9829\\29.9336\\22.4265}&\tabincell{c}{3.6079\\30.7096\\26.3443\\17.1144}\\\hline
   
   DRBN~\cite{yang2020fidelity}&\tabincell{c}{ NIQE\\ PIQE\\ BRISQUE \\ NFERM}&\tabincell{c}{\underline{3.8097}\\58.2841\\27.0778\\19.6805}& \tabincell{c}{3.0884\\49.0913\\28.9623\\17.3524}& \tabincell{c}{3.9699\\47.6355\\31.1219\\24.0582}& \tabincell{c}{2.3226\\46.6456\\28.7414\\19.6118}& \tabincell{c}{3.2033\\47.9635\\32.2557\\24.6923}& \tabincell{c}{3.1183\\49.4907\\29.3885\\19.9499}\\\hline
   
   EnlightenGAN~\cite{jiang2019enlightengan}&\tabincell{c}{ NIQE\\ PIQE\\ BRISQUE \\ NFERM}&\tabincell{c}{{\bf 2.9826}\\\underline{24.0911}\\\underline{18.6199}\\16.0512}& \tabincell{c}{\underline{2.7665}\\28.1605\\22.4538\\\underline{7.6547}}& \tabincell{c}{{\bf 3.3393}\\\underline{33.0809}\\\underline{20.7257}\\21.3110}& \tabincell{c}{{\bf 2.0537}\\\underline{25.7459}\\\underline{18.1831}\\{\bf 11.5793}}& \tabincell{c}{{\bf 2.8636}\\35.6744\\23.2675\\17.2832}& \tabincell{c}{{\bf 2.7089}\\\underline{28.7503}\\\underline{20.9886}\\\underline{12.4296}}\\\hline
   
   Ours&\tabincell{c}{ NIQE\\ PIQE\\ BRISQUE \\ NFERM}&\tabincell{c}{3.8371\\{\bf20.2289}\\{\bf 17.9775}\\{\bf4.2417}}& \tabincell{c}{2.8946\\{\bf21.5900}\\{\bf 19.0040}\\{\bf5.3971}}& \tabincell{c}{4.1777\\{\bf27.0635}\\{\bf 19.5633}\\13.7314}& \tabincell{c}{2.2504\\{\bf21.8966}\\{\bf 17.4819}\\13.8131}& \tabincell{c}{3.3454\\\underline{29.0251}\\28.0440\\12.7063}& \tabincell{c}{\underline{3.0714}\\{\bf23.1712}\\{\bf 20.0525}\\{\bf 8.9968}}\\
   
\bottomrule
\end{tabular*}
\end{table*}

We also conduct quantitative evaluation on all the models via some full-reference and no-reference image quality metrics, including PSNR, SSIM~\cite{wang2004image}, MS-SSIM, NIQE~\cite{mittal2012making}, PIQE~\cite{venkatanath2015blind}, BRISQUE~\cite{mittal2011blind} and NFERM~\cite{gu2014using}. Among them, higher scores of PSNR, SSIM and MS-SSIM indicate better performance. As for the rest metrics, smaller scores of NIQE, PIQE, BRISQUE and NFERM indicate better perceptual quality. Of note, the full-reference image quality metrics are only investigated on the pairwise LOL dataset. Meanwhile, the four no-reference image quality metrics are investigated for all the datasets.

In Table~\ref{tab1}, we report the scores of three full-reference image assessment metrics for all the models. The PSNR/SSIM/MS-SSIM scores of our model are better than those of the other competitors, showing the effectiveness of our method. In Table~\ref{tab2}, we report the scores of four no-reference image quality metrics. On average, EnlightenGAN achieves the best performance in terms of NIQE, while our method obtains the second-best performance in terms of NIQE and the best performances of the other three metrics. On the one hand, our performance on the non-reference image quality metrics again shows that our method can better improve the perceptual quality of an image. On the other hand, by investigating the performances on the DICM/LIME/VV/Fusion datasets, our model still obtains competitive scores, demonstrating that our model trained on the LOL dataset generalizes well to other low-light datasets with various scenes and lightness distributions. The later observation, as well as the visual results in Figs.~\ref{fig6} and~\ref{fig7}, empirically validates the robustness of our model.

\subsubsection{Ablation Studies for Our Model}
We first conduct an ablation study on the model structure by comparing our model with three incomplete versions. The first incomplete version (called Network-\Rmnum{1}) only contains Network-\Rmnum{1}, in which Eq.~\ref{eq3} is taken as the overall loss function. The second one (called Network-\Rmnum{2}(w/o $\mathbf{G}$)) only uses Network-\Rmnum{2} without the branch of the guidance from $\mathbf{G}$, where Eq.~\ref{eq7} works as the loss function. The third one (called Network-\Rmnum{1}+\Rmnum{2}(w/o $\mathbf{G}$)) directly combines Network-\Rmnum{1} and Network-\Rmnum{2} without the $\mathbf{G}$-guidance branch. The third incomplete version adopts Eq.~\ref{eq3} and Eq.~\ref{eq7} as the loss functions for the first and second stage, respectively. It is noted that the second version Network-\Rmnum{2}(w/o $\mathbf{G}$) is trained in an end-to-end way, in which the original low-light images are taken as training inputs and the corresponding normal-light images as the supervision. Besides, since $\mathcal{L}_{r2}$ in Eq.~\ref{eq7} can be seen as the term measuring the lightness difference, Network-\Rmnum{2}(w/o $\mathbf{G}$) has the ability to improve the lightness as well. In this context, Network-\Rmnum{2}(w/o $\mathbf{G}$) can also be regarded as an un-decoupled version, in which the degeneration factors are addressed as a whole. The difference between our model and Network-\Rmnum{2}(w/o $\mathbf{G}$) lies in the decoupled structure of our model. 
\begin{table*}[htbp]
\caption{Comparison of PSNR, SSIM~\cite{wang2004image} and MS-SSIM scores of different network structures}
\label{tab3}
\small
\tabcolsep 3.5pt 
\begin{tabular*}{\textwidth}{c|c|c|c|c}
\toprule
  Metrics &\multicolumn{4}{c}{PSNR/SSIM~\cite{wang2004image}/MS-SSIM} \\\hline
  Model&Network-\Rmnum{1}&Network-\Rmnum{2}(w/o $\mathbf{G}$)&Network-\Rmnum{1}+\Rmnum{2}(w/o $\mathbf{G}$)&Network-\Rmnum{1}+\Rmnum{2}(Ours)\\ \hline
  LOL~\cite{wei2018deep} &17.5083/0.6499/0.8944&\underline{21.1692}/\underline{0.8182}/\underline{ 0.9596}&21.0086/0.8061/0.9493&{\bf 21.8382}/{\bf 0.8216}/{\bf 0.9606} \\
\bottomrule
\end{tabular*}
\end{table*}
\begin{figure*}[htbp]
\centering
\includegraphics[width=\linewidth]{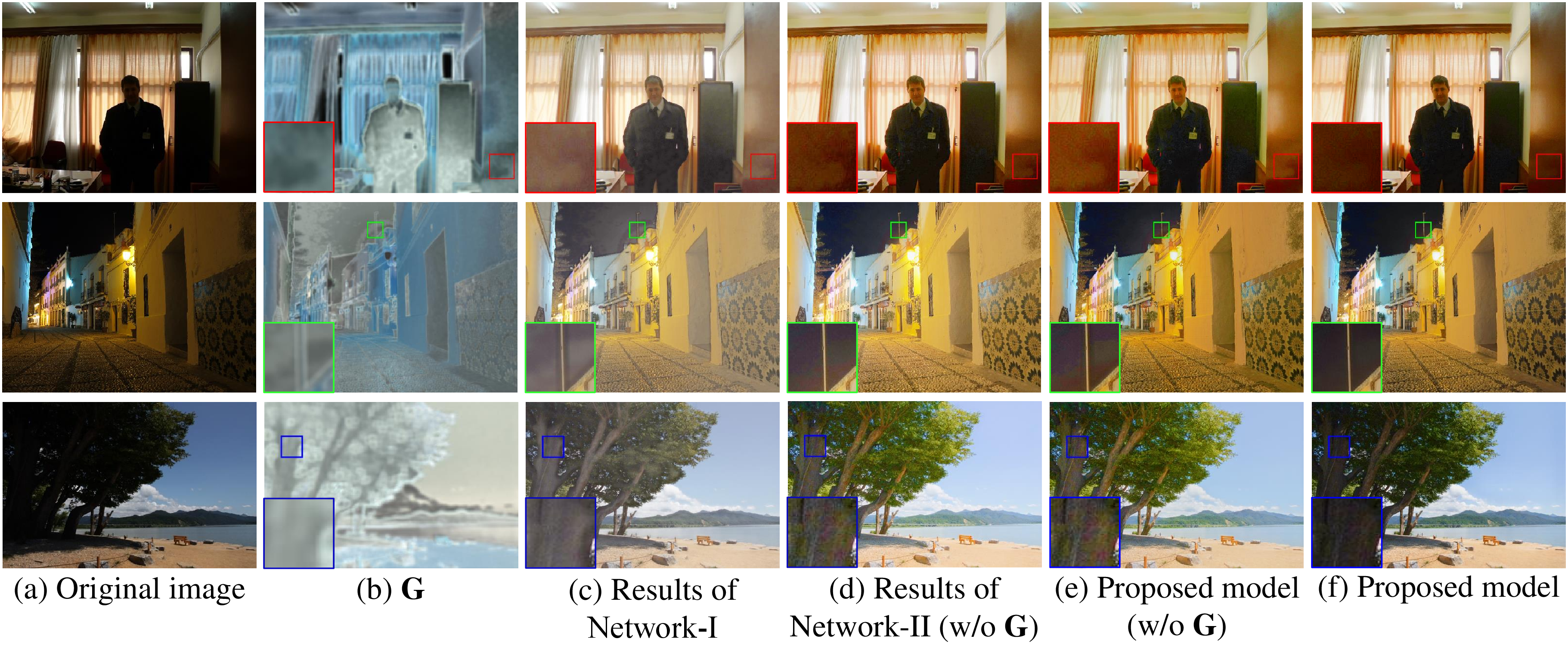}
\caption{Ablation study of the model structure.}
\label{fig8}
\end{figure*}

The PSNR/SSIM/MS-SSIM scores of the three incomplete versions and our model are reported in Table~\ref{tab3}. We also show several examples of enhanced images produced by them in Fig.~\ref{fig8}. In general, we can see that our model is able to process images with different illumination conditions. Furthermore, we have several detailed observations on these results. First, Network-\Rmnum{1} effectively improves the lightness, but the visual quality is still not satisfying, as demonstrated in column (c) of Fig.~\ref{fig8}. Second, by comparing the scores of Network-\Rmnum{1} and Network-\Rmnum{1}+\Rmnum{2}(w/o $\mathbf{G}$) in Table~\ref{tab3}, we can see the quality of the enhanced images has been clearly promoted by tailing Network-\Rmnum{2}(w/o $\mathbf{G}$) after Network-\Rmnum{1}. In Fig.~\ref{fig8}, it is obvious that the visual quality of the images in column (e) is much better than those in column (c). Third, by comparing the third version and our model in Table~\ref{tab3}, we can see that the quantitative metrics are further promoted by utilizing the learned $\mathbf{G}$ map in Network-\Rmnum{2}. By comparing the images in columns (e) and (f) in Fig.~\ref{fig8}, the local appearance of enhanced images is further improved under the guidance of $\mathbf{G}$, making them more natural. The fourth observation comes from the comparison between Network-\Rmnum{2}(w/o $\mathbf{G}$) and Network-\Rmnum{1}+\Rmnum{2}. On the one hand, as shown in Table~\ref{tab3}, our model Network-\Rmnum{1}+\Rmnum{2} performs better than Network-\Rmnum{2}(w/o $\mathbf{G}$) in terms of the quantitative scores. On the other hand, compared with column (d) in Fig.~\ref{fig8}, the results in column (f) better prevent the local over-enhancement, providing more natural visual effects. As for the reason, the decoupled Network-\Rmnum{1}+\Rmnum{2} produces the auxiliary information $\mathbf{G}$ that effectively guides the enhancement task at the second stage, as the learned $\mathbf{G}$ is both aware of the illumination distribution and the content structure of the input image. In all, the above observations empirically validate the effectiveness of each part of our model. 

\begin{table*}[htbp]
\caption{Ablation study on the loss functions, as well as the normalization strategy.}
\label{tab4}
\small
\tabcolsep 10pt 
\begin{tabular*}{\textwidth}{c|c|c|c|c|c|c|c}
\toprule
  Metrics &\multicolumn{7}{c}{PSNR/SSIM~\cite{wang2004image}/MS-SSIM} \\\hline
  Model&w/o $\mathcal{L}_{s}$&w/o $\mathcal{L}_{r2}$&w/o $\mathcal{L}_{vgg}$&w/o $\mathcal{L}_{c}$&w/o LN&w/ BN&Ours\\ \hline
  LOL~\cite{wei2018deep} &\tabincell{c}{21.1504\\0.8113\\\underline{0.9597}}&\tabincell{c}{\underline{21.2414}\\0.7952\\0.9584 
}&\tabincell{c}{20.9045\\0.7982\\0.9250}&\tabincell{c}{20.9106\\0.7700\\0.9582}&\tabincell{c}{20.3886 \\\underline{0.8212}\\0.9508}&\tabincell{c}{17.0974\\0.7193\\0.9238}&\tabincell{c}{{\bf 21.8382}\\{\bf 0.8216}\\{\bf 0.9606}} \\
\bottomrule
\end{tabular*}
\end{table*}
\begin{figure*}[htbp]
\centering
\includegraphics[width=\linewidth]{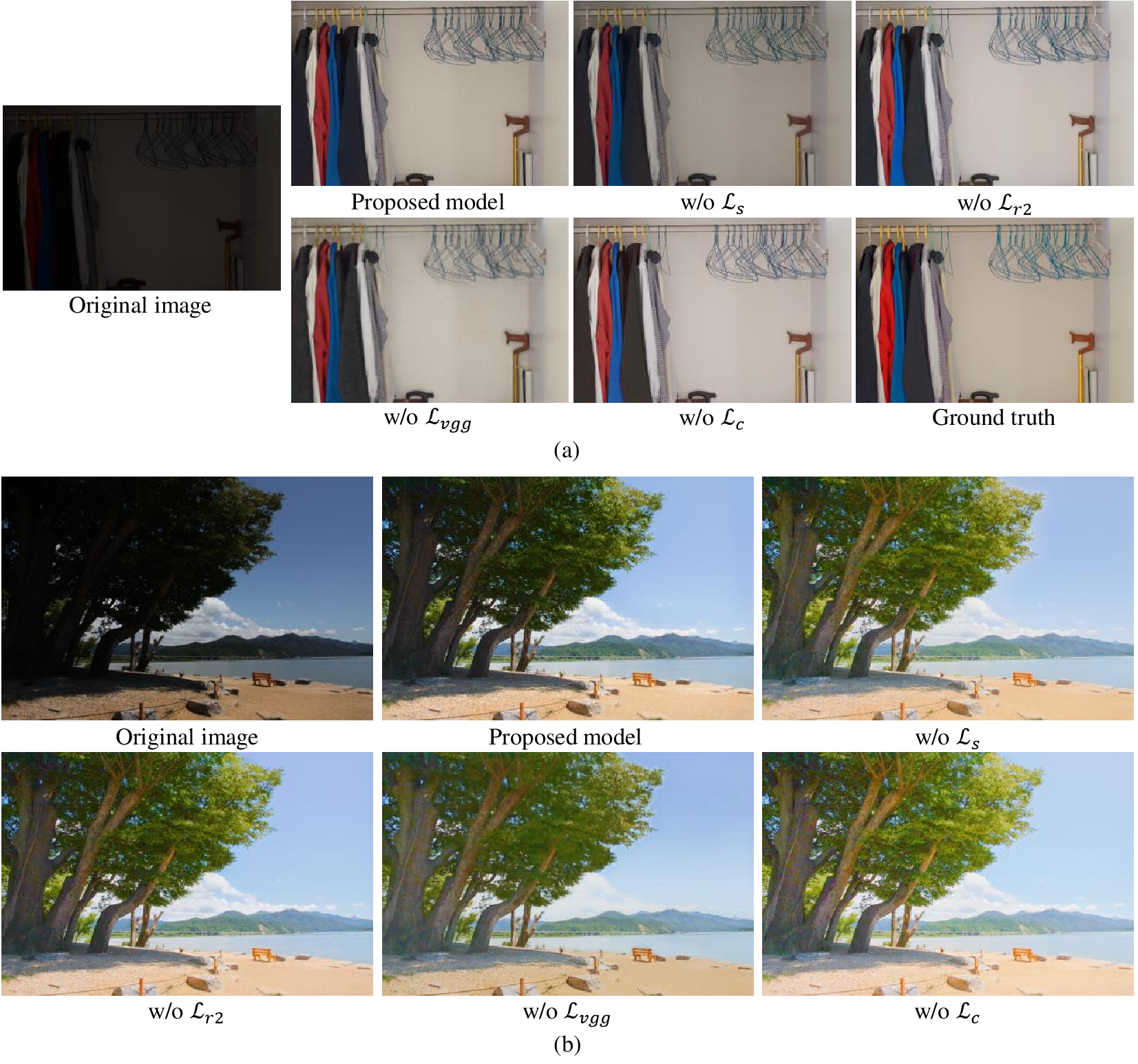}
\caption{Ablation study on the loss function.\iffalse (a) Original image. (b) Proposed model. (c) w/o $\mathcal{L}_{s}$. (d) w/o $\mathcal{L}_{r2}$. (e) w/o $\mathcal{L}_{vgg}$. (f) w/o $\mathcal{L}_{c}$. (h) Target image.\fi}
\label{fig9}
\end{figure*}

In the following, we investigate the effectiveness of each term in the loss functions. In the first four columns of Table~\ref{tab4}, we report the obtained scores based on different loss function configurations. For fair comparison, the structure of our model is fixed as Network-\Rmnum{1}+\Rmnum{2}. By comparing with the scores of the full loss functions, we can see the incremental effectiveness of each loss term. In Fig.~\ref{fig9}, we also provide two visual examples from the LOL test set and the DICM dataset. In Fig.~\ref{fig9} (a), compared with the ground truth image, the imbalanced enhancement in the result of w/o $\mathcal{L}_{s}$ shows the usefulness of the local smoothness imposed on $\mathbf{G}$. The over-enhancement in the result of w/o $\mathcal{L}_{r2}$ shows the necessity of keeping pixel-wise fidelity at the second stage of our model. The unclear local details in the result of w/o $\mathcal{L}_{vgg}$ validate the effectiveness of introducing the perceptual loss. The color distortion in the result of w/o $\mathcal{L}_{c}$ shows the usefulness of introducing the color loss. In Fig.~\ref{fig9} (b), although the ground truth image is not available, we still have the similar observations mentioned above. In addition, the overall naturalness of the proposed model is clearly better than other four incomplete versions.
\begin{figure*}[htbp]
\centering
\includegraphics[width=\linewidth]{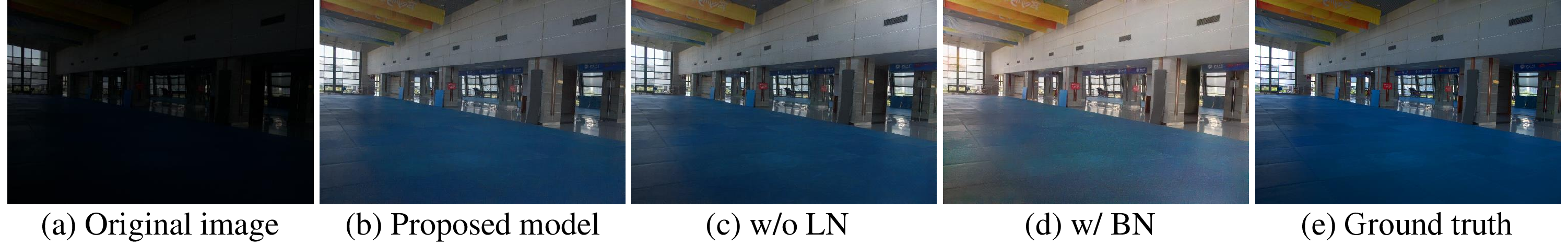}
\caption{Ablation study on the normalization strategy. \iffalse(a) Original image. (b) Proposed model(w/ LN). (c) w/o LN. (c) w/ BN. (d) Target image.\fi}
\label{fig10}
\end{figure*}

We also investigate the effectiveness of the Layer Normalization (LN)~\cite{ba2016layer} technique used in Network-\Rmnum{2}. By comparing the results of w/o LN and ours in Table~\ref{tab4}, we can empirically validate the usefulness of applying LN. In addition, we also replace LN with the Batch Normalization (BN) technique ~\cite{ioffe2015batch}. As shown in Table~\ref{tab4}, the result of w/BN is much worse than ours, which also validates the usefulness of LN. Similar observations can be made from Fig.~\ref{fig10}.

\begin{figure*}[htbp]
\centering
\includegraphics[width=\linewidth]{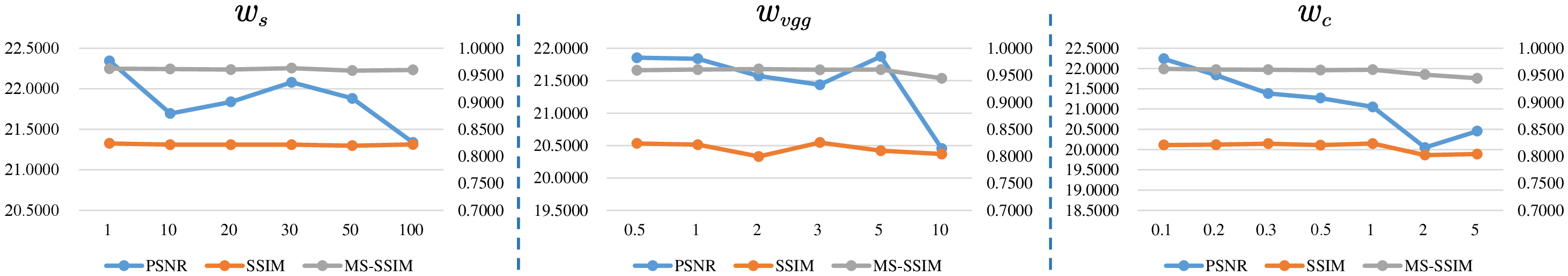}
\caption{The impacts of the weights in Eq.~\ref{eq3} and~\ref{eq7}. These weights are empirically set as $w_{s}=20$, $w_{vgg}=1$ and $w_{c}=0.2$ in all the other experiments.}
\label{fig11}
\end{figure*}

We study the impacts of three weights in Eq.~\ref{eq3} and~\ref{eq7} by varying one of them and fixing the other two. Based on the performance curves of PSNR/SSIM/MS-SSIM in Fig.~\ref{fig11}, we can see that SSIM and MS-SSIM are generally stable with respect to $w_{s}$, $w_{vgg}$ and $w_{c}$ at a wide range. As PSNR only calculates the pixel-wise difference between two images, it has some fluctuations across the ranges of weight value. In spite of this, the overall performance in Fig.~\ref{fig11} still keeps on a competitive level. 

\begin{figure*}[htbp]
\centering
\includegraphics[width=\linewidth]{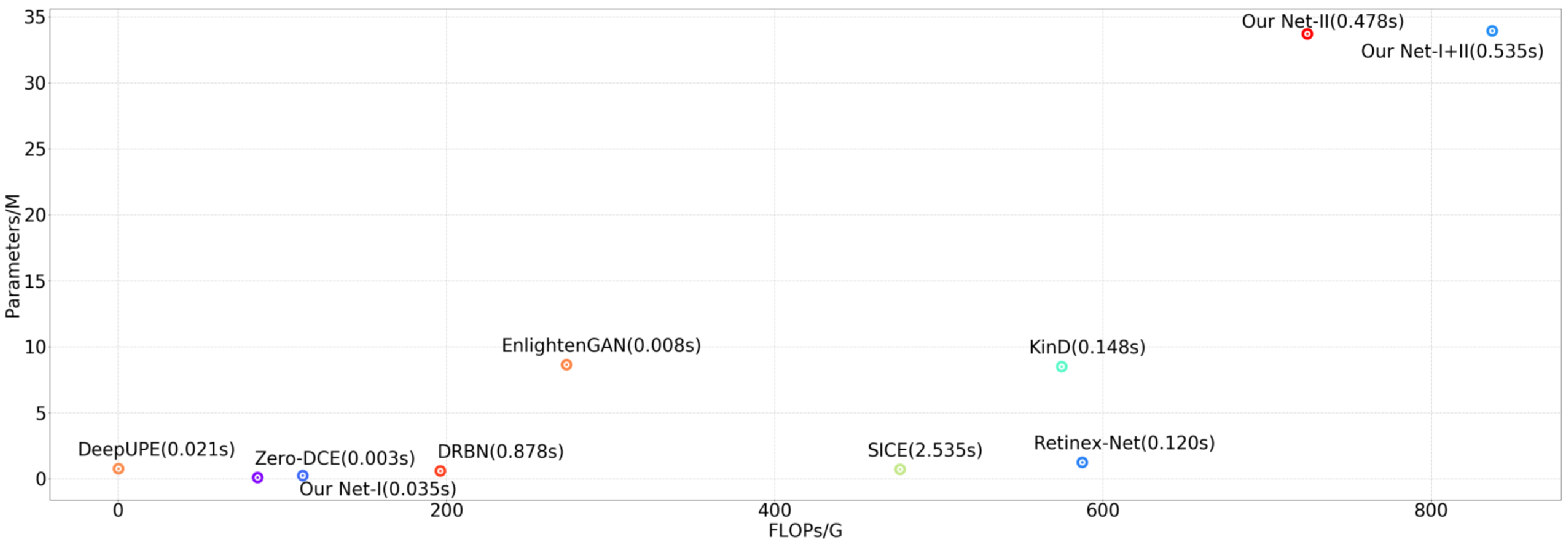}
\caption{ The computational efficiency performance of the eight data-driven methods.}
\label{fig12}
\end{figure*}

At last, we investigate the computational efficiency of the data-driven models used in our experiments. The evaluation metrics are FLOPs (G) and average running time (s) on 1200$\times$900 color images, and the model parameter number (M). Fig.~\ref{fig12} reports the performance of the different models. Our model has the largest values of Parameters and FLOPs, showing that our computational costs are high. As a step further, we can see that the computational costs of Network-\Rmnum{1} are very light (comparable to the other lightweight models), while the costs are mainly from Network-\Rmnum{2} of our model, which aims to finely adjust image appearance and suppress the distortion factors. It is understandable that the large costs are the price of ensuring the overall visual quality of the enhanced results.

\section{Conclusion}
\label{section5}

To obtain satisfying performance, a low-light image enhancement model is expected to suppress multiple degeneration factors other than low lightness, such as imaging noise and color distortion. Current models often focus on enhancing the visibility only, or suppress all the factors as a whole, which possibly lead to sub-optimal results. To solve this issue, in this paper, we build a two-stage model to decouple the low-light enhancement process. Based on the simple pixel-wise non-linear mapping, the first stage only aims to enhance the visibility. Then, the second stage is modeled as the enhancement under normal lightness, in which the remaining degeneration factors are suppressed. In the experiments, qualitative and quantitative comparisons with other models show the superiority of our model. The ablation studies also validate the effectiveness of our model in various aspects.

Our future research includes two directions. First, we can adopt the semi-supervised or the unsupervised strategy to relieve the issue of limited pairwise low/normal-light images needed in fully supervised models. For example, it may be possible to build a semi-supervised model by incorporating the learned $\mathbf{G}$ into some unsupervised model, making the enhancement more aware of the scene structure. Second, we plan to reduce the model size and the computational cost of Network-\Rmnum{2}, making it more compatible with mobile applications.

\begin{acks}
This work was supported in part by the National Nature Science Foundation of China under Grant No. 62172137, 61725203, and 62072152, and in part by the Fundamental Research Funds for the Central Universities under Grant No. PA2020GDKC0023 and PA2019GDZC0095.
\end{acks}
\bibliographystyle{ACM-Reference-Format}
\bibliography{sample-base}

\end{document}